# Evaluation of Multidisciplinary Effects of Artificial Intelligence with Optimization Perspective

*M. Hanefi Calp*
Karadeniz Technical University
Department of Management Information Systems
Faculty of Economics & Administrative Sciences
61080, Trabzon, Turkey
Phone: +90 462 377 3000
mhcalp@ktu.edu.tr

**Abstract**
Artificial Intelligence has an important place in the scientific community as a result of its successful outputs in terms of different fields. In time, the field of Artificial Intelligence has been divided into many sub-fields because of increasing number of different solution approaches, methods, and techniques. Machine Learning has the most remarkable role with its functions to learn from samples from the environment. On the other hand, intelligent optimization done by inspiring from nature and swarms had its own unique scientific literature, with effective solutions provided for optimization problems from different fields. Because intelligent optimization can be applied in different fields effectively, this study aims to provide a general discussion on multidisciplinary effects of Artificial Intelligence by considering its optimization oriented solutions. The study briefly focuses on background of the intelligent optimization briefly and then gives application examples of intelligent optimization from a multidisciplinary perspective.

**Keywords:** Artificial Intelligence; Intelligent Optimization; Optimization; Swarm Intelligence; Metaheuristics.

## 1. Introduction

Technology is a triggering factor in changing life style of people communities and developing the future of the world. After the Industrial Revolution, technological developments have gained a positive momentum to provide effective technological tools and because of that, our life has always affected by new technologies. At this point, computer and communication technologies have also another revolutionary role in especially 20$^{th}$ century and they have caused the virtual world to dominate the real world (Jones, 1998, Winston, 2002, Thurlow et. al., 2004). In detail, there are many different kinds of technological tools discussed widely but Artificial Intelligence and the intelligent technologies are recently has the spotlights on them.

Nowadays, there is a remarkable interest in using intelligent technologies to solve especially daily life problems. The role of Artificial Intelligence in transforming our lives into an automated one and making our daily life more practical in this way has been a usual thing for majority of people. But on the background, there is a long historical way taken by the field of Artificial Intelligence. Although it was firstly affected scientific community and scientific research studies greatly, results of technological developments have started to affect our lives after especially start of the 21$^{st}$ century. Today, it is possible to see effects of Artificial Intelligence and its intelligent approaches, methods, and techniques in almost all fields of the life (Russell & Norvig, 2016; Michalski et. al., 2013; McCorduck, 2009).

Because real world problems are generally associated with classification, suggestion, clustering, control, prediction, optimization or recognition, Artificial Intelligence has become an important solution tool for many fields (Russell & Norvig, 2016; Nilsson, 2014; Nabiyev, 2005; Elmas, 2007; Ghahramani, 2015; Calp & Akcayol, 2018; Dener & Calp, 2018; Sahin et. al., 2014; Erkalan et. al., 2012). Actually, our problems in even different fields of the life can be solved with those solution approaches and because intelligent techniques – algorithms by Artificial Intelligence has enough flexibility and robustness to be adapted in different forms of problems, popularity of Artificial Intelligence has increased rapidly day-by-day. At this point, especially optimization has an effective role on improving the literature of Artificial Intelligence and it has even formed a separate literature in time. Although optimization techniques – algorithms of Artificial Intelligence has no mechanism like learning from samples (Machine Learning), they are widely used in real world problems.

Objective of this study is to provide a general discussion of the role of Artificial Intelligence, by considering its role in solving problems with optimization perspective. Because the life is full of optimization, using intelligent optimization techniques – algorithms should be evaluated by focusing on different fields, which means a multidisciplinary perspective. That research is a general discussion on why the Artificial Intelligence is an effective scientific and technological field and why its optimization oriented solutions can be discussed to understand success of that field better. It is believed that this study will be an effective reference for anyone, who is interested in Artificial Intelligence, intelligent optimization and general applications in this manner.

Associated with the objective of the study, organization of the remaining content is as follows: The next section is devoted to general information about what is optimization and how traditional optimization was transformed into intelligent optimization by Artificial Intelligence. Additionally, the section also provides information about mechanisms that Artificial Intelligence runs in order to achieve intelligent optimization. After that, the third section discussed about multidisciplinary optimization applications of Artificial Intelligence. In this way, it is aimed to enable readers to understand more about how intelligent optimization can be realized in different fields. Following to the third section, a future direction perspective is provided under the fourth section, in order to evaluate the future with current information. Finally, the content is finished with a general discussion on conclusions.

## 2. Optimization Perspective for Artificial Intelligence

In this section, it is aimed to discuss some about how we can evaluate use of Artificial Intelligence from the optimization perspective. First of all, some discussion on what is optimization and how we can see mechanisms of the Artificial Intelligence as an optimization will be provided. After that, especially use of intelligent optimization techniques will be discussed by considering different fields.

### 2.1. From Traditional Optimization to Intelligent Optimization

From a mathematical view, optimization can be defined as finding the most appropriate value (in terms of maximization or minimization) of a variable (or variables) for an objective model (Govan, 2006; Kose 2017). Here, the optimization can be realized with alternative ways but the main point is to find the appropriate value by considering the objective mathematical model, its constants, variables, coefficients, and even constraints. Because the events in the life can be actually defined with mathematical models, it is possible to say that the optimization is associated with all the life and its problems. Because of that, optimization has been one of the most remarkable issues of especially mathematics and statistics (Kose, 2017; Torn & Zilinskas, 1989).

In a traditional manner, optimization is associated with use of some certain mathematical rules in order to solve the optimization task. But as the life becomes more difficult and advanced to be supported with technology, more advanced optimization problems have started to require use of effective and efficient solution ways. Although even advanced optimization problems can be solved finally as a result of some efforts, it has become not a good way to use traditional solution ways when the time and even real-time solution became a vital factor in solution. Because of that, alternative techniques – algorithms have been introduced to the literature. Hill Climbing Algorithm is known as one these techniques and it has been widely used for optimization problems (Ohashi et. al., 2003; Luke, 2009). But because problems of real-life is a typical ocean related to technological improvements, more advanced optimization problems required more advanced solution ways. At this point, Artificial Intelligence has become the most effective technology to derive some solutions for solving even most advanced optimization problems.

Mathematically, an optimization problem can be defined as follows (Kose, 2017; Karaboga, 2014):

$$f(v,y) = cv + y + A \qquad (1)$$

*g(), h(), ...etc.*

Considering the Equation 1, the optimization problem can include objective function(s) like *f(), g(), h(),...etc.* In an objective function, *v,* and *y* corresponds to variable(s), *c* is for coefficient(s), and the function may be supported with constants like *A*. It is important to mention that the variables in the function(s) can be limited with some constraints defined (Kose, 2017; Karaboga, 2014; Mucherino & Seref, 2009). The main objective is to find appropriate value(s) maximizing or minimizing the objective function(s). As the model becomes more complex, it is more vital to use advanced solution ways. Intelligent optimization is currently known as the most effective one in this manner.

Before explaining the mechanisms of intelligent optimization, it is important to explain what kind of optimization tasks are generally solved with intelligent optimization. These are actually associated with the formation of problems but the formation of problems has the role of shaping solution techniques – algorithms in Artificial Intelligence. First of all, the optimization problems done by Artificial Intelligence are classified mainly into two types of optimization: continuous optimization, and combinatorial optimization. They can be defined briefly as follows:

- **Continuous Optimization:** Continuous optimization is associated with finding continuous – real numbers of optimization problems. Here, problems should be modeled with objective function(s) and the function(s) should be supported with the appropriate variables, coefficients, constants, and constraints, in order to define the problem accurately (Kose, 2017; Andréasson et. al., 2005).
- **Combinatorial Optimization:** Combinatorial optimization is focused on finding the most appropriate combination of some known potential solution components, in order to reach to the desired output. In combinatorial optimization, the objective function(s) can be modeled easily but the main point here is to model the potential solution components properly (Papadimitriou & Steiglitz, 1998).
- Except from the related classification, the optimization tasks are also defined with their application time. From that perspective, there are two different optimization ways: static optimization, and dynamic optimization. It is possible to define them as follows:
- **Static Optimization:** Static optimization is done at a certain, specific time and because of that it does not generally need the accurate evaluation of how many times the optimization will be realized and how the time is important in this manner (Leonard et. al., 1992).
- **Dynamic Optimization:** Dynamic optimization is done along a time period so it is important to control the optimization and the problem accurately. Because of that, it is possible to express that the dynamic optimization may be also called as real time optimization (Kamien & Schwartz, 2012).
- Another way of optimization is associated with the number of objective functions considered in the optimization problem. Considering the objective functions, it is possible to talk about single-objective optimization, and multi-objective optimization:
- **Single-objective Optimization:** When the target optimization model has only one objective function like *f()*, that is exactly a single-objective optimization problem (Kose, 2017; Deb, 2014).

- **Multi-objective Optimization:** When the target optimization model consists of more than one objective function like *f(), g(), h(),...etc.*, it is possible to express that this optimization problem is a multi-objective one. Here, functions may be included in some other ones and the problem can be more complex to be solved because of that (Kose, 2017; Deb, 2014).

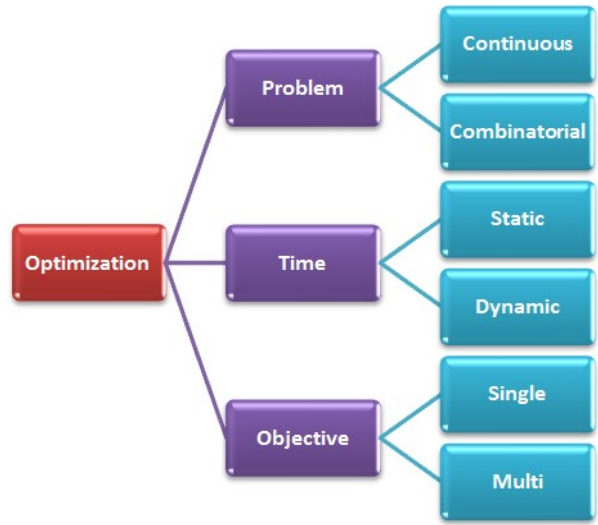

*Figure 1. Classification of optimization from different perspectives*

### 2.2. Intelligent Optimization Mechanisms

In order to achieve intelligent optimization mechanisms, Artificial Intelligence generally takes its power from randomness, chance factor, and heuristic solution approaches. Because the real life is in a flow with lots of randomness and heuristic events caused within some chance factors. Of course, it can be possible to predict the future or instant events if all the variables are known but because that's impossible with current technological state, even intelligent optimization explains that with chance factor. As general, essential intelligent optimization mechanisms used by Artificial Intelligence can be explained briefly as follows:

- **Random Walk:** Randomness is achieved by using random values in intelligent optimization solutions. Use of random values in this manner causes the particles in intelligent optimization techniques – algorithms to move within the solution space. That is briefly called as random walk. In some forms of intelligent optimization solutions, those random walks are sometimes based on some natural movements of swarms (i.e. Levy flights) (Kose, 2017; Karaboga, 2014; Wang, 2001; Reynolds & Rhodes, 2009).
- **Swarm Intelligence:** Intelligent optimization is generally inspired from the nature. In detail, intelligent optimization algorithms employ algorithmic steps including mathematical and logical approaches as connected with natural dynamics or living organisms. When living organisms are considered, it is possible to say that collective behaviors of swarms are important inspirations for intelligent optimization of Artificial Intelligence. Furthermore, swarm oriented techniques – algorithms have caused forming a literature called as Swarm Intelligence under the field of Artificial Intelligence (Kose, 2017; Karaboga, 2014; Kose & Vasant, 2018; Bonabeau et. al., 1999; Eberhart et. al.; 2001; Panigrahi et. al., 2011).
- **Algorithmic Flow:** As general, intelligent optimization techniques are in the form of iterative algorithmic forms, by considering the other mechanisms expressed in this section. As long as the techniques use particles and direct them in the solution space till a stopping criterion, intelligent optimization mechanism can be generally explained with some typical algorithmic flows (Kose, 2017; Wang, 2001).
- **Heuristic and Meta-heuristic:** Intelligent optimization solutions are based on heuristic approaches. In addition to the mentioned other mechanisms so far, that mechanism allows the techniques – algorithms to deal with the optimization tasks. But in time, it has been seen that some heuristic techniques – algorithms are problem dependent and because of that there has been a need for problem-independent solutions. In time, meta-heuristic techniques – algorithms, which are problem-independent, have been introduced to the associated literature (Gendreau & Potvin, 2010; Vasant et. al., 2017; Boussaï et. al., 2013; Yang, 2010; Sörensen et. al., 2018).

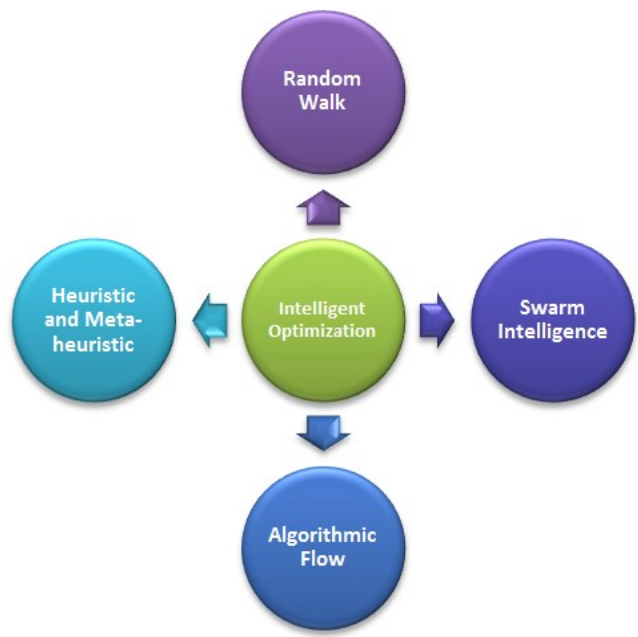

*Figure 2. Intelligent optimization mechanisms used by Artificial Intelligence*

Considering the related intelligent optimization mechanisms, it is possible to see many multidisciplinary applications. Such applications are generally based on continuous or combinatorial optimization problems. Furthermore, some alternative Artificial Intelligence techniques (i.e. Artificial Neural Networks) can be seen as optimization oriented solutions. Eventually, there is a multidisciplinary characteristic of Artificial Intelligence event it is evaluated in the context of optimization approach.

### 3. Multidisciplinary Intelligent Optimization Applications

Multidisciplinary apply of intelligent optimization is a widely followed research interest in the scientific community. Because it is an effective way to formulate real world problems with an optimization oriented approach, different fields are often interested in using Artificial Intelligence and its optimization solutions. Some application examples can be discussed briefly as follows:

- **Design:** One of the most remarkable uses of intelligent optimization is design. The design concept here can be a physical design of a product, technical engineering of a machine or just finding appropriate components of a mathematical model explaining the exact design. In the associated literature, efforts for the optimization done over design problems are called briefly as design optimization. Readers are referred to (Sobieszczanski-Sobieski & Haftka, 1997; Rao et. al., 2011; Gandomi et. al., 2015; Aydogdu et. al., 2016; Baykasoglu & Ozsoydan, 2015; De et. al., 2015; Mirjalili et. al., 2017) for some remarkable and recent applications.
- **Planning:** Planning is an important factor on solving many different types of real world problems. When it is evaluated in terms of optimization, planning may be accepted as a typical combinatorial optimization approach. In the applications, planning can be done for determining a business strategy, movement of logistic vehicles, figuring out paths / routes that buses can follow in a city, or determining an educational strategy / learning path that will be followed at an institution by considering currently rising popularity of interactive and Web based educational solutions (Yang et. al., 2012; Wu et. al., 2010; Zhong et. al., 2018). On the other hand, the optimization may be a continuous one over a function to organize the whole planning. Eventually, it is important to determine exact potential solution components or values and determine the optimum way via intelligent optimization in this way. Some additional, recent and remarkable examples of intelligent optimization for planning can be found at (Chaurasia & Singh, 2015; Kumawat et. al., 2017; Modiri et. al., 2015; Wang et. al., 2016; Fister et. al., 2015; Lemos et. al., 2018).

- **Prediction:** Although the prediction problem is generally solved with Machine Learning techniques, sometimes it can be possible to have some predictive data by using optimization approaches. Here, the prediction problem is tried to be solved via optimization modelling. Because the prediction is important for many fields of the modern life, it is possible to see such applications of intelligent optimization from mechanical engineering to energy studies or from biology / medical oriented studies to data oriented problems of different fields. Readers are referred to (Martinez et. al., 2010; De Carvalho et. al., 2010; Calp, 2019) for some example applications.
- **Control:** Sometimes the problem of control can be effectively solved with optimization oriented solutions. In this way, it can be possible to provide even dynamic control solutions for some real world problems. The control concept here can be also a technical term to meet with the planning, which is more dynamic as occurred real-time. Some examples of using intelligent optimization for control tasks can be examined from (Bhatt et. al., 2010; Kusiak & Zhang, 2010; Connor et. al., 2017) by readers.
- **Optimizing Machine Learning:** Nowadays, a widely followed research way is to use Artificial Intelligence techniques in order to form hybrid systems. In detail, that is done in order to improve effectiveness and accurateness of the solution designed for the objective problem. In this sense, intelligent optimization is often used by researchers in order to train Machine Learning techniques. At this point, because the training of Machine Learning techniques is a typical optimization approach, use of intelligent optimization techniques in this way can be accepted as an application approach. In the literature, some well-known techniques like Artificial Neural Networks or Support Vector Machines are often employed by researchers, in order to have effective processes for training and desired results at the end (Camci et. al., 2018; Aljarah, et. al., 2018; Thom et. al., 2018; Aljarah et. al., 2018; Calp, 2019).

**4. Future Directions**

Explanations so far have focused on how intelligent optimization is achieved thanks to Artificial Intelligence and also how that solution way is done in a multidisciplinary manner. Moving from the explanations and considering the current state of the associated literature, it is also possible to express some perspectives on the future state. In this context, some remarkable predictions on the future are as follows:

- The literature of intelligent optimization is very active and there is a great potential for the future. There has been always a remarkable effort to design and develop new optimization techniques – algorithms and it seems that the future will be same because of the unstoppable nature of arising optimization problems along with the changing, dynamic life – real world.
- Intelligent optimization techniques – algorithms are widely used for solving problems. However, these techniques – algorithms are also used for optimizing other Artificial Intelligence based solutions or traditional, mathematical solutions. In this context, researchers are generally motivated about developing hybrid systems using that solution process. Because the literature of Artificial Intelligence is currently finding its way through solutions, it possible to predict that the future will be connected with increased use of hybrid systems.
- Because of rapid technological developments, the need for dynamic and instant optimization solutions increases in time. At this point, the future will be probably based on using dynamic optimization and developing robust enough techniques – solution ways to deal with that issue.
- Because the real world is surrounded with all continuous and combinatorial optimization oriented problems, popularity of those optimization mechanisms will be increasing in the future. For the continuous optimization, it can be said that the complexity of multi-objective optimization problems will increase and that complexity will affect also modelling combinatorial optimization as the life will always bring more complex form of all optimization solution types.
- It is clear that the multidisciplinary application scope of intelligent optimization will be alive in the future. Because Artificial Intelligence has enough ability to deal with even predicted future problems and it is widely supported with newly developed technologies and scientific fields.
- It is also clear that newer problems of real life will be tried to be solved by Artificial Intelligence and its related techniques. At this point, intelligent optimization will always have a flexible solution scope and employ a strong role in providing desired outputs for us.

## 5. Conclusions

This study provided a general discussion on how intelligent optimization by Artificial Intelligence is realized effectively in a multidisciplinary perspective. In detail, the study has started to explain essentials of optimization and focused how the transformation to intelligent optimization was done. Additionally, the study has also focused on how intelligent optimization mechanisms are realized within designed techniques – algorithms. In this way, it is possible to understand how optimization takes place in the real life and how intelligent optimization has important role for solving the real problems. The literature of intelligent optimization is still remarkable for researchers and has a great potential to improve the Artificial Intelligence literature. The study has also provided some explanations in order point reasons for that situation. Following to that, the study has also provided some discussions on multidisciplinary applications of intelligent optimization and expressed some ideas about future directions.

**References**
Aljarah, I., Ala'M, A. Z., Faris, H., Hassonah, M. A., Mirjalili, S., & Saadeh, H. (2018). Simultaneous feature selection and support vector machine optimization using the grasshopper optimization algorithm. Cognitive Computation, 1-18.
Aljarah, I., Faris, H., & Mirjalili, S. (2018). Optimizing connection weights in neural networks using the whale optimization algorithm. Soft Computing, 22(1), 1-15.
Andréasson, N., Evgrafov, A., Patriksson, M., Gustavsson, E., & Önnheim, M. (2005). An introduction to continuous optimization: foundations and fundamental algorithms (Vol. 28). Lund: Studentlitteratur.
Aydogdu, I., Akin, A., & Saka, M. P. (2016). Design optimization of real world steel space frames using artificial bee colony algorithm with Levy flight distribution. Advances in Engineering Software, 92, 1-14.
Baykasoglu, A., & Ozsoydan, F. B. (2015). Adaptive firefly algorithm with chaos for mechanical design optimization problems. Applied Soft Computing, 36, 152-164.
Bhatt, P., Roy, R., & Ghoshal, S. P. (2010). GA/particle swarm intelligence based optimization of two specific varieties of controller devices applied to two-area multi-units automatic generation control. International journal of electrical power & energy systems, 32(4), 299-310.

Bonabeau, E., Marco, D. D. R. D. F., Dorigo, M., Théraulaz, G., & Theraulaz, G. (1999). Swarm Intelligence: From Natural to Artificial Systems (No. 1). Oxford University Press.

BoussaïD, I., Lepagnot, J., & Siarry, P. (2013). A survey on optimization metaheuristics. Information Sciences, 237, 82-117.

Calp, M. H. (2019). A Hybrid ANFIS-GA Approach for Estimation of Regional Rainfall Amount. Gazi University Journal of Science, 32 (1) (Accepted - In Press).

Calp, M. H. (2019). An estimation of personnel food demand quantity for businesses by using artificial neural networks. Journal of Polytechnic, DOI: 10.2339/politeknik.444380 (Accepted - In Press).

Calp, M. H., & Akcayol, M. A. (2018). Optimization of Project Scheduling Activities in Dynamic CPM and PERT Networks Using Genetic Algorithms. Süleyman Demirel University Journal of Natural and Applied Sciences (SDU J Nat Appl Sci), 22(2), 615-627.

Camci, E., Kripalani, D. R., Ma, L., Kayacan, E., & Khanesar, M. A. (2018). An aerial robot for rice farm quality inspection with type-2 fuzzy neural networks tuned by particle swarm optimization-sliding mode control hybrid algorithm. Swarm and evolutionary computation, 41, 1-8.

Chaurasia, S. N., & Singh, A. (2015). A hybrid swarm intelligence approach to the registration area planning problem. Information Sciences, 302, 50-69.

Connor, J., Seyedmahmoudian, M., & Horan, B. (2017). Using particle swarm optimization for PID optimization for altitude control on a quadrotor. In Universities Power Engineering Conference (AUPEC), 2017 Australasian (pp. 1-6). IEEE.

De Carvalho, A. B., Pozo, A., & Vergilio, S. R. (2010). A symbolic fault-prediction model based on multiobjective particle swarm optimization. Journal of Systems and Software, 83(5), 868-882.

De, B. P., Kar, R., Mandal, D., & Ghoshal, S. P. (2015). Optimal selection of components value for analog active filter design using simplex particle swarm optimization. International Journal of Machine Learning and Cybernetics, 6(4), 621-636.

Deb, K. (2014). Multi-objective optimization. In Search methodologies (pp. 403-449). Springer, Boston, MA.

Degiacomi, M. T., & Dal Peraro, M. (2013). Macromolecular symmetric assembly prediction using swarm intelligence dynamic modeling. Structure, 21(7), 1097-1106.

Dener, M., & Calp, M. H. (2018). Solving the exam scheduling problems in central exams with genetic algorithms. Mugla Journal of Science and Technology, 4(1), 102-115.

Eberhart, R. C., Shi, Y., & Kennedy, J. (2001). Swarm Intelligence. Elsevier.

Elmas, C. (2007). Artificial Intelligence Applications (Artificial Neural Network, Fuzzy Logic, Genetic Algorithms) – (In Turkish). Seckin Press.

Erkalan, M., Calp, M. H., Sahin, I. (2012). Designing and realizing an expert system to be used to profession choosing by utilizing multiple intelligences theory. International Journal of Informatics Technologies. 5(2), 49-55.

Fister, I., Rauter, S., Yang, X. S., & Ljubič, K. (2015). Planning the sports training sessions with the bat algorithm. Neurocomputing, 149, 993-1002.

Gandomi, A. H., Kashani, A. R., Roke, D. A., & Mousavi, M. (2015). Optimization of retaining wall design using recent swarm intelligence techniques. Engineering Structures, 103, 72-84.

Gendreau, M., & Potvin, J. Y. (2010). Handbook of Metaheuristics (Vol. 2). New York: Springer.

Ghahramani, Z. (2015). Probabilistic machine learning and artificial intelligence. Nature, 521(7553), 452.

Govan, A. (2006). Introduction to optimization. In North Carolina State University, SAMSI NDHS, Undergraduate workshop.

Jones, S. (Ed.). (1998). Cybersociety 2.0: Revisiting Computer-mediated Community and Technology (Vol. 2). Sage Publications.

Kamien, M. I., & Schwartz, N. L. (2012). Dynamic Optimization: The Calculus of Variations and Optimal Control in Economics and Management. Courier Corporation.

Karaboga, D. (2014). Artificial Intelligence Optimization Algorithms (In Turkish). Nobel Press.

Kose, U. (2017). Development of Artificial Intelligence Based Optimization Algorithms (In Turkish). PhD. Thesis. Selcuk University, Dept. of Computer Engineering.

Kose, U., & Vasant, P. (2018). A Model of Swarm Intelligence Based Optimization Framework Adjustable According to Problems. In Innovative Computing, Optimization and Its Applications (pp. 21-38). Springer, Cham.

Kumawat, M., Gupta, N., Jain, N., & Bansal, R. C. (2017). Swarm-Intelligence-Based Optimal Planning of Distributed Generators in Distribution Network for Minimizing Energy Loss. Electric Power Components and Systems, 45(6), 589-600.

Kusiak, A., & Zhang, Z. (2011). Adaptive control of a wind turbine with data mining and swarm intelligence. IEEE Transactions on Sustainable Energy, 2(1), 28-36.

Lemos, J. Y., Joshi, A. R., D'souza, M. M., & D'souza, A. D. (2018). Design a Smart and Intelligent Routing Network Using Optimization Techniques. In Data Management, Analytics and Innovation (pp. 297-310). Springer, Singapore.

Leonard, D., Van Long, N., & Ngo, V. L. (1992). Optimal Control Theory and Static Optimization in Economics. Cambridge University Press.

Luke, S. (2009). Essentials of Metaheuristics (Vol. 113). Raleigh: Lulu.

Martinez, E., Alvarez, M. M., & Trevino, V. (2010). Compact cancer biomarkers discovery using a swarm intelligence feature selection algorithm. Computational biology and chemistry, 34(4), 244-250.

McCorduck, P. (2009). Machines Who Think: A Personal Inquiry into the History and Prospects of Artificial Intelligence. AK Peters / CRC Press.

Michalski, R. S., Carbonell, J. G., & Mitchell, T. M. (Eds.). (2013). Machine Learning: An Artificial Intelligence Approach. Springer Science & Business Media.

Mirjalili, S., Gandomi, A. H., Mirjalili, S. Z., Saremi, S., Faris, H., & Mirjalili, S. M. (2017). Salp Swarm Algorithm: A bio-inspired optimizer for engineering design problems. Advances in Engineering Software, 114, 163-191.

Modiri, A., Gu, X., Hagan, A., & Sawant, A. (2015, May). Improved swarm intelligence solution in large scale radiation therapy inverse planning. In Biomedical Conference (GLBC), 2015 IEEE Great Lakes (pp. 1-4). IEEE.

Mucherino, A., & Seref, O. (2009). Modeling and solving real-life global optimization problems with meta-heuristic methods. In Advances in Modeling Agricultural Systems (pp. 403-419). Springer, Boston, MA.

Nabiyev, V. V. (2005). Artificial Intelligence: Problems, Methods, Algorithms (In Turkish). Seckin Press.

Nilsson, N. J. (2014). Principles of Artificial Intelligence. Morgan Kaufmann.

Ohashi, T., Aghbari, Z., & Makinouchi, A. (2003, June). Hill-climbing algorithm for efficient color-based image segmentation. In IASTED International Conference on Signal Processing, Pattern Recognition, and Applications (pp. 17-22).

Panigrahi, B. K., Shi, Y., & Lim, M. H. (Eds.). (2011). Handbook of Swarm Intelligence: Concepts, Principles and Applications (Vol. 8). Springer Science & Business Media.

Papadimitriou, C. H., & Steiglitz, K. (1998). Combinatorial Optimization: Algorithms and Complexity. Courier Corporation.

Rao, R. V., Savsani, V. J., & Vakharia, D. P. (2011). Teaching–learning-based optimization: a novel method for constrained mechanical design optimization problems. Computer-Aided Design, 43(3), 303-315.

Reynolds, A. M., & Rhodes, C. J. (2009). The Lévy flight paradigm: random search patterns and mechanisms. Ecology, 90(4), 877-887.

Russell, S. J., & Norvig, P. (2016). Artificial Intelligence: A Modern Approach. Malaysia; Pearson Education Limited,.

Sahin, I., Calp, M. H., & Ozkan, A. (2014). An expert system design and application for hydroponics greenhouse systems. Gazi University Journal of Science, 27(2).

Sobieszczanski-Sobieski, J., & Haftka, R. T. (1997). Multidisciplinary aerospace design optimization: survey of recent developments. Structural optimization, 14(1), 1-23.

Sörensen, K., Sevaux, M., & Glover, F. (2018). A history of metaheuristics. Handbook of Heuristics, 1-18.

Thom, H. T., Ming-Yuan, C. H. O., & Tuan, V. Q. (2018). A novel perturbed particle swarm optimization-based support vector machine for fault diagnosis in power distribution systems. Turkish Journal of Electrical Engineering & Computer Sciences, 26(1), 518-529.

Thurlow, C., Lengel, L., & Tomic, A. (2004). Computer Mediated Communication. Sage.

Torn, A., & Zilinskas, A. (1989). Global Optimization. Springer-Verlag New York, Inc..

Vasant, P., Kose, U., & Watada, J. (2017). Metaheuristic techniques in enhancing the efficiency and performance of thermo-electric cooling devices. Energies, 10(11), 1703.

Wang, G. G., Chu, H. E., & Mirjalili, S. (2016). Three-dimensional path planning for UCAV using an improved bat algorithm. Aerospace Science and Technology, 49, 231-238.

Wang, L. (2001). Intelligent Optimization Algorithms with Applications. Tsinghua University & Springer Press, Beijing.

Winston, B. (2002). Media, Technology and Society: A History: From the Telegraph to the Internet. Routledge.

Wu, Z., Ni, Z., Gu, L., & Liu, X. (2010). A revised discrete particle swarm optimization for cloud workflow scheduling. In Computational Intelligence and Security (CIS), 2010 International Conference on (pp. 184-188). IEEE.

Yang, X. S. (2010). Nature-inspired Metaheuristic Algorithms. Luniver Press.

Yang, X. S., Deb, S., Karamanoglu, M., & He, X. (2012). Cuckoo search for business optimization applications. In Computing and Communication Systems (NCCCS), 2012 National Conference on (pp. 1-5). IEEE.

Zhong, S., Zhou, L., Ma, S., Jia, N., Zhang, L., & Yao, B. (2018). The optimization of bus rapid transit route based on an improved particle swarm optimization. Transportation Letters, 10(5), 257-268.

**Hanefi M. Calp** received his Ph. D. degree in 2018 from the department of Management Information Systems at Gazi University, one of the most prestigious universities in Turkey. He works as Assistant Professor in the department of Management Information Systems of the Faculty of Economics & Administrative Sciences of the Karadeniz Technical University. His research interest includes Management Information Systems, Artifical Neural Networks, Expert Systems, Fuzzy Logic, Risk Management, Risk Analysis, Human-Computer Interaction, Technology Management, Knowledge Management, Digital Transformation and Project Management.

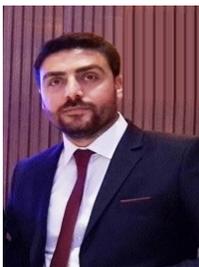